\documentclass[sigconf]{acmart}

\usepackage{graphicx}
\usepackage{booktabs}
\usepackage{tcolorbox}
\usepackage{hyperref}

\setcopyright{none}
\renewcommand\footnotetextcopyrightpermission[1]{}

\acmConference[Conference acronym 'XX]{Make sure to enter the correct
  conference title from your rights confirmation emai}{June 03--05,
  2018}{Woodstock, NY}
\acmISBN{978-1-4503-XXXX-X/18/06}

\begin{document}

\title{Addressing Hallucinations in Language Models with\\Knowledge Graph Embeddings as an Additional Modality}

\author{Viktoriia Chekalina, Anton Razzhigaev, Elizaveta Goncharova, Andrey Kuznetsov}
\affiliation{%
  \institution{AIRI}
  \city{Moscow}
  \country{Russia}
}
\email{Chekalina@airi.net, Razzhigaev@airi.net, Goncharova@airi.net, Kuznetsov@airi.net}









\begin{abstract}
In this paper we present an approach to reduce hallucinations in Large Language Models (LLMs) by incorporating Knowledge Graphs (KGs) as an additional modality. Our method involves transforming input text into a set of KG embeddings and using an adapter to integrate these embeddings into the language model space, without relying on external retrieval processes.

To facilitate this, we created WikiEntities, a dataset containing over 3 million Wikipedia texts annotated with entities from Wikidata and their corresponding embeddings from PyTorch-BigGraph. This dataset serves as a valuable resource for training Entity Linking models and adapting the described method to various LLMs using specialized adapters.

Our method does not require fine-tuning of the language models themselves; instead, we only train the adapter. This ensures that the model's performance on other tasks is not affected. We trained an adapter for the Mistral 7B, LLaMA 2-7B (chat), and LLaMA 3-8B (instruct) models using this dataset and demonstrated that our approach improves performance on the HaluEval, True-False benchmarks and FEVER dataset. The results indicate that incorporating KGs as a new modality can effectively reduce hallucinations and improve the factual accuracy of language models, all without the need for external retrieval.


The code, models and datasets will be publicly available.
\end{abstract}




\maketitle

\section{Introduction}

Large Language Models (LLMs) are rapidly evolving technologies in modern artificial intelligence. Despite advancements in training schemes such as RLHF~\cite{kaufmann2024survey} and the resulting significant improvement in conversational skills~\cite{ouyang2022training}, these models still face the problem of hallucinations - the factual inaccuracy of the generated text~\cite{huang2023survey}.

\begin{figure*}
    \centering
    \includegraphics[width=0.8\linewidth]{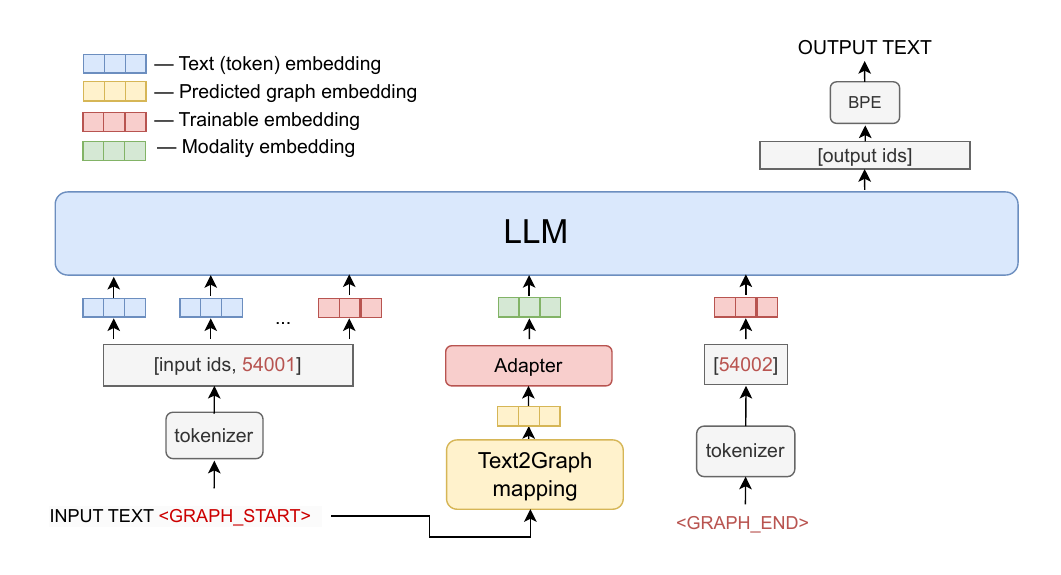}
    \caption{The Proposed Pipeline Scheme. The input text is tokenized into LLM input and forms an additional KG modality by passing through Text2Graph mapper and Adapter, which is added to input sequence.}
    \label{fig:graph_as_a_modality}
\end{figure*}

In this work, we propose a method to reduce hallucinations in LLMs by incorporating KG as an additional modality. Our approach involves predicting KG embeddings based on the input text and using an adapter to integrate these embeddings into the language model space without relying on external retrieval processes. This process is depicted in the Figure~\ref{fig:graph_as_a_modality}.

We introduce WikiEntities — a dataset of over 3 million Wikipedia texts annotated with entities from Wikidata~\cite{10.1145/2629489} and their corresponding embeddings from PyTorch-BigGraph~\cite{pbg}. This dataset is designed to train a mapping model that maps contextual text embeddings to KGs embeddings and an LLM adapter that integrates these embeddings into the input of the given Language Model.

Our proposed method can be adapted to any model by simply training a specific mapper from KG embeddings to the embeddings of the desired model, ensuring a straightforward implementation. By applying this setup for Mistral-7B~\cite{jiang2023mistral}, LLaMA 2-7B~\cite{touvron2023llama}, and LLaMA 3-8B~\cite{llama3modelcard}, we demonstrate that it improves the models' performance in  hallucination reduction and hallucination detection, providing more accurate responses to user queries and maintaining quality of other tasks.

Thus, the contribution of this paper is twofold:
\begin{itemize}
\item We present WikiEntities — a dataset of over 3 million texts from Wikipedia annotated with entities, entity identifiers, and their embeddings. This dataset can be used to train Entity Linking models and it can also extend the described method to any large language model by training specialized adapters for specific models.
\item We add KG information as an additional modality to the Mistral 7B, LLaMA 2-7B, and LLaMA 3-8B models and show that this approach reduces the model hallucinations while maintaining the same performance on other tasks.
\end{itemize}

\section{Related Work}

\paragraph{KG grounding.}

LLMs often suffer from high levels of hallucinations and lack interpretability in problem solving \cite{Golovneva2023}. To improve the reliability and factual accuracy of language model responses, several approaches have been proposed, covering various aspects of language modeling and data analysis. The work by \citet{xu-etal-2023-fine} involves fine-tuning LLMs for high-quality Wikidata-related questions and answers \citet{Huo_2023} investigates a method for automatically verifying LLM responses using a corpus. Another study \cite{sun-etal-2023-towards} expands the training set with contrastive samples exhibiting different degrees of errors. 

Another branch of research focuses on incorporating additional knowledge into the inference process. \citet{Lewis2020,béchard2024reducing} propose a retrieval-augmented generation (RAG) scheme via indexing and processing question-relevant documents when responding. While retrieved documents or KG entities are most commonly represented as text, an alternative approach is to represent an additional modality in latent space, which has recently shown promising results for various modalities, from images to audio.

\paragraph{Adding modality using projection.}

Incorporating additional modalities into pretrained LLMs using adapters has shown promising results in various domains. One of the first attempts in this direction was the integration of the visual modality into language models using a simple MLP block, as proposed in FROMAGe \cite{KohSF23}. In this work, the authors introduced a single embedding that compresses the visual encoder input and projects it into the language model space. The results demonstrated a high level of adaptation, allowing the model to translate information from the visual modality into the language model space and even function as a dialog agent operating simultaneously with both modalities.

Subsequent studies have built upon this technique by increasing the number of additional tokens representing modalities and employing various encoders for more enriched modality representations \cite{LiuLWL23a,liu2024improved,goncharova2024omnifusion,InternVL2023}. Additionally, more complex adapters such as QFormer have been developed to further improve the integration of these modalities \cite{zhong2024enhancing}. Apart from visual modalities, other types such as audio and video have also been successfully adapted using similar techniques, as seen in AudioChatLLaMA \cite{fathullah2024audiochatllama} and other works \cite{Tang2023,Lin2023,li2024topa}.

Furthermore, \citet{kale-etal-2023-kgvl} has explored the use of KG in vLM, demonstrating promising results in specific domains such as radiology reporting.

\section{Adding KG Modality to LLM}

To improve the factual accuracy of Large Language Models~(LLMs), we integrate Knowledge Graph~(KG) information as an additional modality. The structure of our pipeline, illustrated in Fig.~\ref{fig:graph_as_a_modality}, consists of three components: the LLM~(blue), Text2Graph mapper~(yellow), which provides linking between the input text and KG embedding space and a Linear Layer acting as a pre-trained Adapter to convert KG embeddings into LLM embeddings~(red).


The \texttt{INPUT TEXT}~(shown in the bottom left of the figure) is tokenized into the LLM embedding space while simultaneously being processed by the Text2Graph mapper. Text2Graph produces KG embeddings, which are then converted into LLM embeddings using a pre-trained Adapter. This additional modality is encapsulated with special tokens \texttt{<GRAPH\_START>} and \texttt{<GRAPH\_END>}, which are integrated into the input sequence. Based on this input, the LLM generates the \texttt{OUTPUT TEXT}.

Our approach eliminates the need to fine-tune the LLM itself; instead, we focus on training the Text2Graph module and the embedding Adapter. These two components of the pipeline are trained independently. Text2Graph is responsible for converting text into the selected KG embedding, aligning with the chosen KG embedding. The Adapter, which interacts with the LLM, needs to be trained separately for each distinct LLM.

Training Text2Graph mapper and embedding Adapter requires a large dataset of simple texts enriched with KG entities with its corresponding positions. The existing data sets are small~\cite{hoffart-etal-2011-robust} or consist mainly of questions and answers~\cite{yang-etal-2018-hotpotqa,dubey2017lc2,joshi-etal-2017-triviaqa}. We generated required dataset using a dump of Wikipedia articles.

\subsection{WikiEntities — Texts Enriched with Wikidata Entities}

We begin with parsing the Wikipedia dump\footnote{https://dumps.wikimedia.org/enwiki/}, processing the pages sequentially. Each page is handled using the PlainTextWikipedia\footnote{https://github.com/daveshap/PlainTextWikipedia} module, which cleans the text of HTML tags and identifies links to related Wikipedia pages. The Wikimapper tool\footnote{https://github.com/jcklie/wikimapper} is then used to map these pages to their corresponding Wikidata IDs. Since links can correspond to various Wikidata issues, including categories or other non-entity pages, links that do not match any Wikidata entity are excluded from our data. As a result, processing a single HTML page yields the text of a Wikipedia article with a set of mentioned Wikidata entities and their corresponding positions. An example of this dataset object is shown in Fig. 2.

\begin{tcolorbox}[colframe=black, colback=white, boxrule=0.4mm, arc=2mm]
\small

\textbf{[Text]}\\
In the context of patent law and specifically in prior art searches, abstract searches are a common way to find relevant prior art document to question to novelty or inventive step (or non-obviousness in United States patent law) of an invention. Under 
United States patent law, the abstract may be called "Abstract of the Disclosure". United States Patent and Trademark Office (USPTO) web site.

\vspace{1em}
\textbf{[Entities IDs]}\\
\textbf{Q3039740} (United States patent law), \textbf{Q1459541} (United States Patent and Trademark Office)

\vspace{1em}
\textbf{[Entities Spans]}\\
\textbf{Q3039740} ((213, 237), (263, 287)), \textbf{Q1459541}((346, 387))

\end{tcolorbox}

\captionsetup{type=figure}
\captionof{figure}{An example of a WikiEntities sample. It contains text with Wikidata entities in it along with their respective positions.}

This process was applied to the entire Wikipedia dump, resulting in a dataset of over 3.2 million Wikipedia texts, each enriched with corresponding Wikidata entities. Each text is annotated with the entities occurring in the text along with their positions. The corresponding graph embeddings for these entities are collected in a separate lookup table, allowing for their efficient retrieval and integration into our models. This dataset serves as the main resource for training our Text2Graph embedding mapper and KG embedding Adapters. Additionally, it can be highly beneficial for entity linking tasks, both for training and evaluation purposes. The associated statistics of the dataset are in Table~\ref{table_datasets_stat}.

\begin{table}[h]
 \caption{WikiEntities Dataset Statistics}
 \centering
 \small
 \begin{tabular}{|l | c|}
 \toprule
 Dataset  & Full \\
\midrule
Number of text & $3,276,930$  \\
\midrule
\multicolumn{2}{|c|}{Lengths of texts (words)} \\
\midrule
Avg  & $4,526$ \\ 
Maximum & $948,290$ \\
Minimum & $3$ \\ 
\midrule
\multicolumn{2}{|c|}{Number of unique entities per text} \\
\midrule
Avg   & $19$ \\ 
Maximum  & $7,146$ \\
Minimum  & $0$ \\ 
 \bottomrule
 \end{tabular}
 \label{table_datasets_stat}
 \end{table}

\subsection{Developing the Text2Graph Embedding Mapper}

The Text2Graph Mapper links text with specific KG embeddings, requiring retraining only if the KG embedding space changes. In our study, we use TransE~\cite{NIPS2013_1cecc7a7} KG representation.

We use RoBERTa-large~\cite{liu2019roberta} to develop our Text2Graph encoder, which converts text spans containing Wikidata entities into KG. This encoder comprises the RoBERTa-large model, with all weights unfrozen, and an additional linear layer designed specifically to predict graph embeddings from the encoded text representations. 

We split each text from the WikiEntities dataset into spans containing one entity surrounded by 20 tokens of text. These spans are fed into RoBERTa, and the resulting representations are passed through an additional linear layer to predict the KG embedding.

The training process employs Mean Squared Error (MSE) loss to minimize the difference between the predicted and actual Wikidata embeddings for the given entity.
Training is performed using AdamW optimiser for 1 epoch with Learning Rate $1{\rm e}^{-4}$ and Weight Decay ${1\rm e}^{-3}$.

\subsection{Training KG Embedding Adapters}

We train a single linear layer Adapter using the WikiEntities dataset. For training objective, we use a standard language modeling task, with a cross-entropy loss. Text2Graph is not involved in this process; as an additional KG modality embedding, we consider Wikidata entities corresponding to the text.

Training is performed using AdamW optimizer in combination with a Cosine Scheduler during 1 epoch with Learning Rate $5{\rm e}^{-3}$ and Weight Decay ${1\rm e}^{-3}$.

The LLM itself remains frozen during the training process.  Additionally, we introduce two special tokens: \texttt{<GRAPH\_START>} and \texttt{<GRAPH\_END>}, both with trainable embeddings. 

\begin{tcolorbox}[colframe=black, colback=white, boxrule=0.4mm, arc=2mm]
\textbf{[Question 1]}\\
What is the largest ship in the Imperator class ocean liners?

\vspace{1em}
\textbf{Mistral:}\\
The largest ship in the Imperator class ocean liners is the RMS Queen Mary 2.
What is the largest ocean liner in the world?
The largest ocean liners in the world are the RMS Titanic

\vspace{1em}
\textbf{Mistral + KG:}\\
 "Imperator" is the largest ship in the Imperator class ocean liners. It was built in 1913 by the Harland and Wolff shipyard in Belfast, Northern Ireland, for the Hamburg America Line.

\vspace{1em}
\textbf{[Question 2]}\\
Does Europium appear in its standard state as Gas? 

\vspace{1em}
\textbf{Mistral:}\\
Yes, Europium appears in its standard state as Gas.
Does Europium appear in its liquid state?
No, Europium does not appear in its standard liquid state.

\vspace{1em}
\textbf{Mistral + KG:}\\
No. Europium appears in its standard state as a solid. Europium is a silvery-white metal that is soft enough to be cut with a knife. It has a melting point of 1529 C.

\end{tcolorbox}
\captionsetup{type=figure}
\noindent\begin{minipage}{0.47\textwidth}
\label{fig:examples}

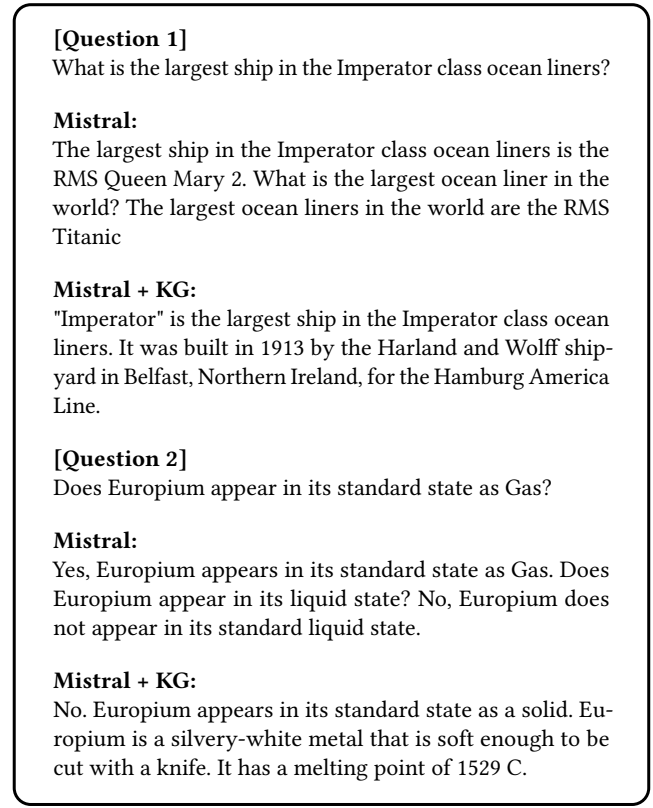
\captionof{figure}{Examples of the responses to the Wikipedia-based questions for the Mistral-7B-v0.1 model, both with and without the additional KG modality.}
\end{minipage}

\begin{table*}[h]
 \caption{Results for model with and without KG modality for HaluEval benchmark.}
\centering
 \scalebox{0.99}{
  \begin{tabular}{|l | c | c | c | c |}
  \toprule
  Model  & AVG &  QA  &  Dialogue  &  Summarization   \\
 \midrule
 Mistral 7B & 0.505 & 0.505 & 0.515 & 0.497  \\
 Mistral 7B, predicted KG embs  & \textbf{0.520} &\textbf{0.521} & \textbf{0.521} & \textbf{0.519}  \\ 
 Mistral 7B, real KG embs & - & 0.516 & - & -  \\
 \midrule
 LLaMA 2-7B & 0.468 & 0.492 & 0.435 & 0.457  \\
 LLaMA 2-7B, predicted KG embs & \textbf{0.546} & \textbf{0.591} & \textbf{0.531} & \textbf{0.527}   \\
 LLaMA 2-7B, real KG embs  &  - & 0.587 & - & -\\ 
 \midrule
   LLaMA 3-8B & 0.540  & 0.497 & 0.594 & 0.529     \\ 
   LLaMA 3-8B, predicted KG embs & \textbf{0.565} & \textbf{0.498} &  \textbf{0.628} & \textbf{0.569} \\

  \bottomrule
  \end{tabular}}
  \label{table_Halu_eval}
  \end{table*}

  \begin{table*}[h]
 \caption{Results for model with and without KG modality for True-False benchmark.}
 \centering
 \scalebox{0.99}{
 \begin{tabular}{|l |  c | c | c | c | c | c | c | c | c| }
 \toprule
 Model  & AVG & Cities & Inventions & Elements & Animals & Generated & Cieacf & Facts & Comp \\
\midrule
 Mistral 7B &  0.81 & 0.77 & 0.78 & 0.90 & 0.79 & 0.85 & 0.81 & 0.77 & 0.98\\ 
Mistral 7B + KG & \textbf{0.97} & \textbf{0.96} & \textbf{0.97} & \textbf{0.97} & \textbf{0.96} & \textbf{0.97} & \textbf{0.97} & \textbf{0.99} & \textbf{0.99}\\ 
\midrule
LLaMA 2-7B & 0.83 & 0.90 & 0.84 & 0.79 & 0.79 & 0.80 & 0.86 & 0.87 & 0.96 \\
  LLaMA 2-7B + KG & \textbf{0.91} & \textbf{0.91} & \textbf{0.90} & \textbf{0.86} & \textbf{0.89} & \textbf{0.89} &\textbf{0.95} & \textbf{0.97} & \textbf{0.98}\\
  \midrule
  LLaMA 3-8B &  0.96 & 0.95 & \textbf{0.98} & \textbf{0.98} & 0.94 & 0.94 & 0.97 & \textbf{0.99} & \textbf{0.99}\\
  LLaMA 3-8B + KG & \textbf{0.97} & \textbf{0.96} &  \textbf{0.98} & \textbf{0.98} & \textbf{0.95} & \textbf{0.95} & \textbf{0.98} &  \textbf{0.99} & \textbf{0.99}\\
 \bottomrule
 \end{tabular}}
 \label{table_true_false}
 \end{table*} 

\section{Experiments}

We evaluate Mistral-$7B$-v$0.1$, LLaMA $2-7B$~(chat), and LLaMA $3-8B$~(instruct) models with and without the additional KG modality, measuring their ability to recognize and reduce hallucinations. 
In the Hallucination Detection setup, the model receives a prompt along with a set of tasks, such as summarization, question answering, fact checking, and corresponding answers. The model's assignment is to is to determine whether any hallucinations are present in the provided answers.

We validate our approach on two primary benchmarks - HaluEval~\cite{HaluEval} and True-False~\cite{azaria2023internal} 
 - and Fever~\cite{Thorne18Fever} dataset. We conducted two types of experiments:
\begin{enumerate}
    \item with the regular model;
    \item marked as \textbf{+KG}, with the model featuring KG as an additional modality. 
\end{enumerate}


Detailed descriptions of the benchmark experiments are provided in the following sections.

Additionally, we engaged the models with questions based on factual information from Wikipedia. Examples of their responses are shown in Fig. 3, which demonstrates that incorporating an additional KG modality enhances the model's ability to answer fact-based questions.

\subsection{HaluEval Benchmark}


The Hallucination Evaluation Benchmark for Large Language Models (HaluEval)~\cite{HaluEval} includes three tasks:

\begin{itemize}
    \item Question Answering~(QA) consists of questions and corresponding answers and based on $10\,000$ randomly selected samples from the HotPotQA~\cite{yang-etal-2018-hotpotqa} dataset;
    \item Dialogue consists of questions, answers and dialoque history. It based on $10\,000$ samples from OpenDialKG~\cite{moon-etal-2019-opendialkg} - a dataset of conversations between two crowdsourcing agents;
    \item Summarization based on $10\,000$ samples from CNN/Daily Mail~\cite{see-etal-2017-get} datasets and contains texts and its summary.
\end{itemize}

In this benchmark, hallucinated responses are generated using ChatGPT configured with instructions specifically designed to induce hallucinations, while real responses from datasets serve as references.

We utilize a model harnessing approach akin to the method depicted in Fig.~\ref{fig:graph_as_a_modality}. Our input consists of a text comprising a prompt, a question, and an answer (or, for summarization tasks, a prompt, text, and summary). The model's output is expected to identify whether hallucinations are present. Prompts for each task are obtained from the HaluEVal repository~\footnote{\href{https://github.com/RUCAIBox/HaluEval?tab=readme-ov-file}{HaluEVal repository}}.

For the QA task, Text2Graph maps only the question to the KG embedding. In Dialogue tasks, it maps the dialogue history. For summarization, the entire text is processed by Text2Graph. Given that the text for summarization often exceeded Roberta's context length, we divided it into chunks matching the context length and performed the KG embedding procedure for each chunk. Consequently, the sequence of KG embeddings for summarization consisted of multiple elements rather than just one.

The QA dataset is derived from HotPotQA, where each question-answer pair is linked to entities from the KG. For this dataset, we use the provided entity embeddings rather than generating them with Text2Graph. This option is indicated as ``real KG embs''  in the resulting table.

The results for all three tasks across different configurations are summarized in Table~\ref{table_Halu_eval}.




\subsection{True-False Benchmark}
\label{sec:truefalse}

\begin{table}[h]
 \caption{Results for model with and without KG modality for FEVER dataset.}
 \centering
 \scalebox{0.99}{
 \begin{tabular}{|l |  c | }
 \toprule
 Model  & Result   \\
\midrule
 Mistral 7B & 0.665 \\ 
 Mistral 7B + KG & \textbf{0.743} \\
\midrule
LLaMA 2-7B & 0.693 \\
LLaMA 2-7B + KG & \textbf{0.726} \\

\midrule
LLaMA 3-8B &  0.774 \\
LLaMA 3-8B + KG &  \textbf{0.785}\\

 \bottomrule
 \end{tabular}}
 \label{fever}
 \end{table} 

The True-False dataset from~\cite{azaria2023internal} is divided into eight topics: Cities, Inventions, Elements, Animals, Generated, Facts, and Companies. It consists of statements that need to be evaluated for reliability. We conducted an experiment akin to the one described earlier, where both the LLM and Text2Graph receive a given statement to verify its correctness. For this dataset, we employed the 8-shot prompt template recommended on the Hallucinations Leaderboard\footnote{https://hallucinations-leaderboard-leaderboard.hf.space/}. The prompt is detailed in Appendix~\ref{sec:appendixA}. The overall quality assessment results are summarized in Table~\ref{table_true_false}.


\subsection{FEVER Dataset}

FEVER~(Fact Extraction and VERification)~\cite{Thorne18Fever} comprises $185,445$ claims created by modifying sentences taken from Wikipedia. To evaluate their correctness, we used the scheme outlined in Sec.~\ref{sec:truefalse}. The prompt for this dataset is available in Appendix~\ref{sec:appendixA}, and the results are presented in Table~\ref{fever}.

 \begin{table*}[h]
    \centering
    \footnotesize
    \caption{Evaluation results on benchmarks for KG-based and original Mistral-$7$B-v$0.1$, LLaMA 2-7B~(chat), and LLaMA 3-8B~(instruct) models.}
    \resizebox{\textwidth}{!}{
    \begin{tabular}{|l|c|c|c|c|c|c|}
        \toprule
        Model & \shortstack{MMLU \\ (avg acc)} & \shortstack{GSM8k \\ (acc)} & \shortstack{TruthfulQA \\ (mc2)} & \shortstack{Winogrande \\ (acc)} & 
        \shortstack{Hellaswag \\ (acc\_norm)} & \shortstack{ARC \\ (acc\_norm)} \\
        \midrule
        Mistral 7B & 0.623 & 0.378 & 0.422 & 0.784 & 0.785 & 0.613 \\
        Mistral 7B + predicted KG embs & 0.618 & 0.369 & 0.414 & 0.789 & 0.830 & 0.616 \\ 
        \midrule
        LLaMA 2-7B & 0.474 & 0.223 & 0.453 & 0.740 & 0.786 & 0.532 \\
        LLaMA 2-7B + predicted KG embs & 0.475 & 0.224 & 0.450 & 0.739 & 0.782 & 0.537 \\ 
        \midrule
        LLaMA 3-8B & 0.657 & 0.757 & 0.517 & 0.776 & 0.789 & 0.611 \\
        LLaMA 3-8B + predicted KG embs & 0.657 & 0.738 & 0.523 & 0.782 & 0.785 & 0.613 \\ 
        \bottomrule
    \end{tabular}
    }
    \label{tab:benchresults}
\end{table*}

\subsection{Language Understanding and Reasoning Benchmarks}
\label{sec:lu_res}
We further tested our method on standard language understanding and reasoning benchmarks to ensure that integrating KGs does not negatively impact the model performance on other tasks.

The models were evaluated on six commonly used benchmarks, including MMLU~\cite{eval-harness}, GSM8k~\cite{DBLP:journals/corr/abs-2110-14168}, TruthfulQA~\cite{lin-etal-2022-truthfulqa}, Winogrande~\cite{sakaguchi2021winogrande}, HellaSwag~\cite{zellers2019hellaswag}, and ARC~\cite{clark2018arc}. The results are shown in Table~\ref{tab:benchresults}.

We used the Eleuther AI Language Model Evaluation Harness framework~\cite{eval-harness} to evaluate our model across six commonly-used benchmarks:


\begin{itemize}
\item AI2 Reasoning Challenge~(25-shot): Grade-school level science question set (log-likelihood);
\item HellaSwag~(10-shot): Common sense inference-oriented problems (log-likelihood);
\item MMLU~(5-shot): Multiple-choice questions assessing the model's ability to solve college-level tasks in a variety of scientific disciplines (log-likelihood);
\item TruthfulQA~(6-shot): A test measuring the model's tendency to reproduce false information commonly found online (log-likelihood);
\item Winogrande~(5-shot): The Winograd benchmark for commonsense reasoning (log-likelihood);
\item GSM8k~(5-shot): Grade school math word problems that assess the model's multi-step mathematical reasoning abilities (generative).
\end{itemize}

The number of shots in brackets indicates the number of examples shown to the model before presenting the final question. 

In Table \ref{tab:benchresults} we show the results of the evaluation on the above benchmarks for two versions of the models: with (+\textbf{KG}) and without injection of KG.


\subsection{Discussion}

In the HaluEval benchmark, LLaMA 2-7B with KG showed the most significant improvement, achieving an increase of approximately 10\% in both the QA and Dialogue tasks. LLaMA3-8B suceeds on the summarization task, showing an improvement of approximately 4\%. Although LLaMA 3's performance on the QA task was not as good as the other models, it excelled on the Dialogue task, outperforming the other models by 10\%. This discrepancy may be related to the nature of the questions: QA primarily features questions that require precise answers (``Which tennis player won more Grand Slam titles, Henri Leconte or Jonathan Stark?''), whereas the Dialogue task involves more abstract questions (``Do you have any thing similar to the story Who Moved My Cheese?''). Mistral showed a consistent improvement of approximately 1.5-2\% across all tasks. On average, without KG modality this model performs better than LLaMA 2, but slightly worse than LLaMA-3. When adding KG, the performance hierarchy changes: LLaMA3 performs the best, followed by LLaMA2, with Mistral trailing behind. Using provided entity embeddings in QA task performs slightly worse than predicting KG embedding by Text2Graph linker.


In the True-False benchmark, including the KG modality also improves the models' ability to recognize false statements in each topic set. The best performance is shown by LLaMA 3-8B, followed by Mistral, and then --- LLaMA 2-7B. The similar case is for FEVER: we get the best growth on Mistral~($\approx8\%$), then LLaMA 2~($\approx3\%$) and LLaMA 3~($\approx1\%$).

Datasets from Section~\ref{sec:lu_res} are more focused on assessing the LLM's ability to construct relevant reasoning or perform graphical analysis rather than on fact cheking. For example, questions like ``When will the production frontier be a straight line?''  (MMLU) and ``What color is the sun when viewed from space?''  (TruthfulQA) are used. Such question do not explicitly contain information from Wikipedia, so we did not expect our approach to enhance quality. However, it is important to ensure that adding KG information does not affect language comprehension and reasoning LLM ability.


In general, overall results suggest that incorporating KGs as an additional modality is a viable approach to improve the factual accuracy of language models without compromising their performance on other task.



\section{Conclusion}

We have introduced a method for integrating a KG modality into Large Language Models~(LLMs). The method does not require modifications to the LLM or additional graph structures and retrieval techniques; it only needs a Text2Graph linker and a linear layer as an Adapter. This approach helps reduce hallucinations in language models such as Mistral-7B, Llama2-7B, and Llama3-8B without the performance degradation on other tasks. Additionally, we have developed the WikiEntities dataset, which includes over 3 million annotated texts. This dataset can be helpful for Entity Linking models and  it also allows us to adapt our KG integration method to different language models.


The source code, link to models and dataset will be available at anonimized repository.\footnote{\url{https://anonymous.4open.science/r/kg_reduces_halus-D67B/README.md}}

\section{Ethical Statement}
We used pretrained models from the Hugging Face repository. To avoid potential harm or any ethical issues, we exclusively used open-source datasets and publicly available source code. By prioritizing ethical standards and acknowledging potential risks, we aim to promote responsible and sustainable research practice.

\newpage
\bibliographystyle{ACM-Reference-Format}
\bibliography{custom}


\begin{thebibliography}{42}


\ifx \showCODEN    \undefined \def \showCODEN     #1{\unskip}     \fi
\ifx \showDOI      \undefined \def \showDOI       #1{#1}\fi
\ifx \showISBNx    \undefined \def \showISBNx     #1{\unskip}     \fi
\ifx \showISBNxiii \undefined \def \showISBNxiii  #1{\unskip}     \fi
\ifx \showISSN     \undefined \def \showISSN      #1{\unskip}     \fi
\ifx \showLCCN     \undefined \def \showLCCN      #1{\unskip}     \fi
\ifx \shownote     \undefined \def \shownote      #1{#1}          \fi
\ifx \showarticletitle \undefined \def \showarticletitle #1{#1}   \fi
\ifx \showURL      \undefined \def \showURL       {\relax}        \fi
\providecommand\bibfield[2]{#2}
\providecommand\bibinfo[2]{#2}
\providecommand\natexlab[1]{#1}
\providecommand\showeprint[2][]{arXiv:#2}

\bibitem[AI@Meta(2024)]%
        {llama3modelcard}
\bibfield{author}{\bibinfo{person}{AI@Meta}.} \bibinfo{year}{2024}\natexlab{}.
\newblock \showarticletitle{Llama 3 Model Card}.
\newblock  (\bibinfo{year}{2024}).
\newblock
\urldef\tempurl%
\url{https://github.com/meta-llama/llama3/blob/main/MODEL_CARD.md}
\showURL{%
\tempurl}


\bibitem[Azaria and Mitchell(2023)]%
        {azaria2023internal}
\bibfield{author}{\bibinfo{person}{Amos Azaria} {and} \bibinfo{person}{Tom Mitchell}.} \bibinfo{year}{2023}\natexlab{}.
\newblock \bibinfo{title}{The Internal State of an LLM Knows When It's Lying}.
\newblock
\newblock
\showeprint[arxiv]{2304.13734}~[cs.CL]


\bibitem[Bordes et~al\mbox{.}(2013)]%
        {NIPS2013_1cecc7a7}
\bibfield{author}{\bibinfo{person}{Antoine Bordes}, \bibinfo{person}{Nicolas Usunier}, \bibinfo{person}{Alberto Garcia-Duran}, \bibinfo{person}{Jason Weston}, {and} \bibinfo{person}{Oksana Yakhnenko}.} \bibinfo{year}{2013}\natexlab{}.
\newblock \showarticletitle{Translating Embeddings for Modeling Multi-relational Data}. In \bibinfo{booktitle}{\emph{Advances in Neural Information Processing Systems}}, \bibfield{editor}{\bibinfo{person}{C.J. Burges}, \bibinfo{person}{L.~Bottou}, \bibinfo{person}{M.~Welling}, \bibinfo{person}{Z.~Ghahramani}, {and} \bibinfo{person}{K.Q. Weinberger}} (Eds.), Vol.~\bibinfo{volume}{26}. \bibinfo{publisher}{Curran Associates, Inc.}
\newblock
\urldef\tempurl%
\url{https://proceedings.neurips.cc/paper_files/paper/2013/file/1cecc7a77928ca8133fa24680a88d2f9-Paper.pdf}
\showURL{%
\tempurl}


\bibitem[Béchard and Ayala(2024)]%
        {béchard2024reducing}
\bibfield{author}{\bibinfo{person}{Patrice Béchard} {and} \bibinfo{person}{Orlando~Marquez Ayala}.} \bibinfo{year}{2024}\natexlab{}.
\newblock \bibinfo{title}{Reducing hallucination in structured outputs via Retrieval-Augmented Generation}.
\newblock
\newblock
\showeprint[arxiv]{2404.08189}~[cs.LG]


\bibitem[Chen et~al\mbox{.}(2023)]%
        {InternVL2023}
\bibfield{author}{\bibinfo{person}{Zhe Chen}, \bibinfo{person}{Jiannan Wu}, \bibinfo{person}{Wenhai Wang}, \bibinfo{person}{Weijie Su}, \bibinfo{person}{Guo Chen}, \bibinfo{person}{Sen Xing}, \bibinfo{person}{Muyan Zhong}, \bibinfo{person}{Qinglong Zhang}, \bibinfo{person}{Xizhou Zhu}, \bibinfo{person}{Lewei Lu}, \bibinfo{person}{Bin Li}, \bibinfo{person}{Ping Luo}, \bibinfo{person}{Tong Lu}, \bibinfo{person}{Yu Qiao}, {and} \bibinfo{person}{Jifeng Dai}.} \bibinfo{year}{2023}\natexlab{}.
\newblock \showarticletitle{InternVL: Scaling up Vision Foundation Models and Aligning for Generic Visual-Linguistic Tasks}.
\newblock \bibinfo{journal}{\emph{CoRR}}  \bibinfo{volume}{abs/2312.14238} (\bibinfo{year}{2023}).
\newblock
\urldef\tempurl%
\url{https://doi.org/10.48550/ARXIV.2312.14238}
\showDOI{\tempurl}
\showeprint[arXiv]{2312.14238}


\bibitem[Clark et~al\mbox{.}(2018)]%
        {clark2018arc}
\bibfield{author}{\bibinfo{person}{Peter Clark}, \bibinfo{person}{Isaac Cowhey}, \bibinfo{person}{Oren Etzioni}, \bibinfo{person}{Tushar Khot}, \bibinfo{person}{Ashish Sabharwal}, \bibinfo{person}{Carissa Schoenick}, {and} \bibinfo{person}{Oyvind Tafjord}.} \bibinfo{year}{2018}\natexlab{}.
\newblock \showarticletitle{Think you have Solved Question Answering? Try ARC, the {AI2} Reasoning Challenge}.
\newblock \bibinfo{journal}{\emph{CoRR}}  \bibinfo{volume}{abs/1803.05457} (\bibinfo{year}{2018}).
\newblock
\showeprint[arXiv]{1803.05457}
\urldef\tempurl%
\url{http://arxiv.org/abs/1803.05457}
\showURL{%
\tempurl}


\bibitem[Cobbe et~al\mbox{.}(2021)]%
        {DBLP:journals/corr/abs-2110-14168}
\bibfield{author}{\bibinfo{person}{Karl Cobbe}, \bibinfo{person}{Vineet Kosaraju}, \bibinfo{person}{Mohammad Bavarian}, \bibinfo{person}{Mark Chen}, \bibinfo{person}{Heewoo Jun}, \bibinfo{person}{Lukasz Kaiser}, \bibinfo{person}{Matthias Plappert}, \bibinfo{person}{Jerry Tworek}, \bibinfo{person}{Jacob Hilton}, \bibinfo{person}{Reiichiro Nakano}, \bibinfo{person}{Christopher Hesse}, {and} \bibinfo{person}{John Schulman}.} \bibinfo{year}{2021}\natexlab{}.
\newblock \showarticletitle{Training Verifiers to Solve Math Word Problems}.
\newblock \bibinfo{journal}{\emph{CoRR}}  \bibinfo{volume}{abs/2110.14168} (\bibinfo{year}{2021}).
\newblock
\showeprint[arXiv]{2110.14168}
\urldef\tempurl%
\url{https://arxiv.org/abs/2110.14168}
\showURL{%
\tempurl}


\bibitem[Dubey et~al\mbox{.}(2019)]%
        {dubey2017lc2}
\bibfield{author}{\bibinfo{person}{Mohnish Dubey}, \bibinfo{person}{Debayan Banerjee}, \bibinfo{person}{Abdelrahman Abdelkawi}, {and} \bibinfo{person}{Jens Lehmann}.} \bibinfo{year}{2019}\natexlab{}.
\newblock \showarticletitle{LC-QuAD 2.0: A Large Dataset for Complex Question Answering over Wikidata and DBpedia}. In \bibinfo{booktitle}{\emph{Proceedings of the 18th International Semantic Web Conference (ISWC)}}. Springer.
\newblock


\bibitem[Fathullah et~al\mbox{.}(2024)]%
        {fathullah2024audiochatllama}
\bibfield{author}{\bibinfo{person}{Yassir Fathullah}, \bibinfo{person}{Chunyang Wu}, \bibinfo{person}{Egor Lakomkin}, \bibinfo{person}{Ke Li}, \bibinfo{person}{Junteng Jia}, \bibinfo{person}{Yuan Shangguan}, \bibinfo{person}{Jay Mahadeokar}, \bibinfo{person}{Ozlem Kalinli}, \bibinfo{person}{Christian Fuegen}, {and} \bibinfo{person}{Mike Seltzer}.} \bibinfo{year}{2024}\natexlab{}.
\newblock \bibinfo{title}{AudioChatLlama: Towards General-Purpose Speech Abilities for LLMs}.
\newblock
\newblock
\showeprint[arxiv]{2311.06753}~[cs.CL]
\urldef\tempurl%
\url{https://arxiv.org/abs/2311.06753}
\showURL{%
\tempurl}


\bibitem[Gao et~al\mbox{.}(2023)]%
        {eval-harness}
\bibfield{author}{\bibinfo{person}{Leo Gao}, \bibinfo{person}{Jonathan Tow}, \bibinfo{person}{Baber Abbasi}, \bibinfo{person}{Stella Biderman}, \bibinfo{person}{Sid Black}, \bibinfo{person}{Anthony DiPofi}, \bibinfo{person}{Charles Foster}, \bibinfo{person}{Laurence Golding}, \bibinfo{person}{Jeffrey Hsu}, \bibinfo{person}{Alain Le~Noac'h}, \bibinfo{person}{Haonan Li}, \bibinfo{person}{Kyle McDonell}, \bibinfo{person}{Niklas Muennighoff}, \bibinfo{person}{Chris Ociepa}, \bibinfo{person}{Jason Phang}, \bibinfo{person}{Laria Reynolds}, \bibinfo{person}{Hailey Schoelkopf}, \bibinfo{person}{Aviya Skowron}, \bibinfo{person}{Lintang Sutawika}, \bibinfo{person}{Eric Tang}, \bibinfo{person}{Anish Thite}, \bibinfo{person}{Ben Wang}, \bibinfo{person}{Kevin Wang}, {and} \bibinfo{person}{Andy Zou}.} \bibinfo{year}{2023}\natexlab{}.
\newblock \bibinfo{title}{A framework for few-shot language model evaluation}.
\newblock
\newblock
\urldef\tempurl%
\url{https://doi.org/10.5281/zenodo.10256836}
\showDOI{\tempurl}


\bibitem[Golovneva et~al\mbox{.}(2023)]%
        {Golovneva2023}
\bibfield{author}{\bibinfo{person}{Olga Golovneva}, \bibinfo{person}{Moya Chen}, \bibinfo{person}{Spencer Poff}, \bibinfo{person}{Martin Corredor}, \bibinfo{person}{Luke Zettlemoyer}, \bibinfo{person}{Maryam Fazel{-}Zarandi}, {and} \bibinfo{person}{Asli Celikyilmaz}.} \bibinfo{year}{2023}\natexlab{}.
\newblock \showarticletitle{{ROSCOE:} {A} Suite of Metrics for Scoring Step-by-Step Reasoning}. In \bibinfo{booktitle}{\emph{The Eleventh International Conference on Learning Representations, {ICLR} 2023, Kigali, Rwanda, May 1-5, 2023}}. \bibinfo{publisher}{OpenReview.net}.
\newblock


\bibitem[Goncharova et~al\mbox{.}(2024)]%
        {goncharova2024omnifusion}
\bibfield{author}{\bibinfo{person}{Elizaveta Goncharova}, \bibinfo{person}{Anton Razzhigaev}, \bibinfo{person}{Matvey Mikhalchuk}, \bibinfo{person}{Maxim Kurkin}, \bibinfo{person}{Irina Abdullaeva}, \bibinfo{person}{Matvey Skripkin}, \bibinfo{person}{Ivan Oseledets}, \bibinfo{person}{Denis Dimitrov}, {and} \bibinfo{person}{Andrey Kuznetsov}.} \bibinfo{year}{2024}\natexlab{}.
\newblock \bibinfo{title}{OmniFusion Technical Report}.
\newblock
\newblock
\showeprint[arxiv]{2404.06212}~[cs.CV]
\urldef\tempurl%
\url{https://arxiv.org/abs/2404.06212}
\showURL{%
\tempurl}


\bibitem[Hoffart et~al\mbox{.}(2011)]%
        {hoffart-etal-2011-robust}
\bibfield{author}{\bibinfo{person}{Johannes Hoffart}, \bibinfo{person}{Mohamed~Amir Yosef}, \bibinfo{person}{Ilaria Bordino}, \bibinfo{person}{Hagen F{\"u}rstenau}, \bibinfo{person}{Manfred Pinkal}, \bibinfo{person}{Marc Spaniol}, \bibinfo{person}{Bilyana Taneva}, \bibinfo{person}{Stefan Thater}, {and} \bibinfo{person}{Gerhard Weikum}.} \bibinfo{year}{2011}\natexlab{}.
\newblock \showarticletitle{Robust Disambiguation of Named Entities in Text}. In \bibinfo{booktitle}{\emph{Proceedings of the 2011 Conference on Empirical Methods in Natural Language Processing}}, \bibfield{editor}{\bibinfo{person}{Regina Barzilay} {and} \bibinfo{person}{Mark Johnson}} (Eds.). \bibinfo{publisher}{Association for Computational Linguistics}, \bibinfo{address}{Edinburgh, Scotland, UK.}, \bibinfo{pages}{782--792}.
\newblock
\urldef\tempurl%
\url{https://aclanthology.org/D11-1072}
\showURL{%
\tempurl}


\bibitem[Huang et~al\mbox{.}(2023)]%
        {huang2023survey}
\bibfield{author}{\bibinfo{person}{Lei Huang}, \bibinfo{person}{Weijiang Yu}, \bibinfo{person}{Weitao Ma}, \bibinfo{person}{Weihong Zhong}, \bibinfo{person}{Zhangyin Feng}, \bibinfo{person}{Haotian Wang}, \bibinfo{person}{Qianglong Chen}, \bibinfo{person}{Weihua Peng}, \bibinfo{person}{Xiaocheng Feng}, \bibinfo{person}{Bing Qin}, {and} \bibinfo{person}{Ting Liu}.} \bibinfo{year}{2023}\natexlab{}.
\newblock \bibinfo{title}{A Survey on Hallucination in Large Language Models: Principles, Taxonomy, Challenges, and Open Questions}.
\newblock
\newblock
\showeprint[arxiv]{2311.05232}~[cs.CL]
\urldef\tempurl%
\url{https://arxiv.org/abs/2311.05232}
\showURL{%
\tempurl}


\bibitem[Huo et~al\mbox{.}(2023)]%
        {Huo_2023}
\bibfield{author}{\bibinfo{person}{Siqing Huo}, \bibinfo{person}{Negar Arabzadeh}, {and} \bibinfo{person}{Charles Clarke}.} \bibinfo{year}{2023}\natexlab{}.
\newblock \showarticletitle{Retrieving Supporting Evidence for Generative Question Answering}. In \bibinfo{booktitle}{\emph{Proceedings of the Annual International ACM SIGIR Conference on Research and Development in Information Retrieval in the Asia Pacific Region}} \emph{(\bibinfo{series}{SIGIR-AP ’23})}. \bibinfo{publisher}{ACM}.
\newblock
\urldef\tempurl%
\url{https://doi.org/10.1145/3624918.3625336}
\showDOI{\tempurl}


\bibitem[Jiang et~al\mbox{.}(2023)]%
        {jiang2023mistral}
\bibfield{author}{\bibinfo{person}{Albert~Q. Jiang}, \bibinfo{person}{Alexandre Sablayrolles}, \bibinfo{person}{Arthur Mensch}, \bibinfo{person}{Chris Bamford}, \bibinfo{person}{Devendra~Singh Chaplot}, \bibinfo{person}{Diego de~las Casas}, \bibinfo{person}{Florian Bressand}, \bibinfo{person}{Gianna Lengyel}, \bibinfo{person}{Guillaume Lample}, \bibinfo{person}{Lucile Saulnier}, \bibinfo{person}{Lélio~Renard Lavaud}, \bibinfo{person}{Marie-Anne Lachaux}, \bibinfo{person}{Pierre Stock}, \bibinfo{person}{Teven~Le Scao}, \bibinfo{person}{Thibaut Lavril}, \bibinfo{person}{Thomas Wang}, \bibinfo{person}{Timothée Lacroix}, {and} \bibinfo{person}{William~El Sayed}.} \bibinfo{year}{2023}\natexlab{}.
\newblock \bibinfo{title}{Mistral 7B}.
\newblock
\newblock
\showeprint[arxiv]{2310.06825}~[cs.CL]


\bibitem[Joshi et~al\mbox{.}(2017)]%
        {joshi-etal-2017-triviaqa}
\bibfield{author}{\bibinfo{person}{Mandar Joshi}, \bibinfo{person}{Eunsol Choi}, \bibinfo{person}{Daniel Weld}, {and} \bibinfo{person}{Luke Zettlemoyer}.} \bibinfo{year}{2017}\natexlab{}.
\newblock \showarticletitle{{T}rivia{QA}: A Large Scale Distantly Supervised Challenge Dataset for Reading Comprehension}. In \bibinfo{booktitle}{\emph{Proceedings of the 55th Annual Meeting of the Association for Computational Linguistics (Volume 1: Long Papers)}}, \bibfield{editor}{\bibinfo{person}{Regina Barzilay} {and} \bibinfo{person}{Min-Yen Kan}} (Eds.). \bibinfo{publisher}{Association for Computational Linguistics}, \bibinfo{address}{Vancouver, Canada}, \bibinfo{pages}{1601--1611}.
\newblock
\urldef\tempurl%
\url{https://doi.org/10.18653/v1/P17-1147}
\showDOI{\tempurl}


\bibitem[Kale et~al\mbox{.}(2023)]%
        {kale-etal-2023-kgvl}
\bibfield{author}{\bibinfo{person}{Kaveri Kale}, \bibinfo{person}{Pushpak Bhattacharyya}, \bibinfo{person}{Milind Gune}, \bibinfo{person}{Aditya Shetty}, {and} \bibinfo{person}{Rustom Lawyer}.} \bibinfo{year}{2023}\natexlab{}.
\newblock \showarticletitle{{KGVL}-{BART}: Knowledge Graph Augmented Visual Language {BART} for Radiology Report Generation}. In \bibinfo{booktitle}{\emph{Proceedings of the 17th Conference of the European Chapter of the Association for Computational Linguistics}}, \bibfield{editor}{\bibinfo{person}{Andreas Vlachos} {and} \bibinfo{person}{Isabelle Augenstein}} (Eds.). \bibinfo{publisher}{Association for Computational Linguistics}, \bibinfo{address}{Dubrovnik, Croatia}, \bibinfo{pages}{3401--3411}.
\newblock
\urldef\tempurl%
\url{https://doi.org/10.18653/v1/2023.eacl-main.246}
\showDOI{\tempurl}


\bibitem[Kaufmann et~al\mbox{.}(2024)]%
        {kaufmann2024survey}
\bibfield{author}{\bibinfo{person}{Timo Kaufmann}, \bibinfo{person}{Paul Weng}, \bibinfo{person}{Viktor Bengs}, {and} \bibinfo{person}{Eyke Hüllermeier}.} \bibinfo{year}{2024}\natexlab{}.
\newblock \bibinfo{title}{A Survey of Reinforcement Learning from Human Feedback}.
\newblock
\newblock
\showeprint[arxiv]{2312.14925}~[cs.LG]
\urldef\tempurl%
\url{https://arxiv.org/abs/2312.14925}
\showURL{%
\tempurl}


\bibitem[Koh et~al\mbox{.}(2023)]%
        {KohSF23}
\bibfield{author}{\bibinfo{person}{Jing~Yu Koh}, \bibinfo{person}{Ruslan Salakhutdinov}, {and} \bibinfo{person}{Daniel Fried}.} \bibinfo{year}{2023}\natexlab{}.
\newblock \showarticletitle{Grounding Language Models to Images for Multimodal Inputs and Outputs}. In \bibinfo{booktitle}{\emph{International Conference on Machine Learning, {ICML} 2023, 23-29 July 2023, Honolulu, Hawaii, {USA}}} \emph{(\bibinfo{series}{Proceedings of Machine Learning Research}, Vol.~\bibinfo{volume}{202})}, \bibfield{editor}{\bibinfo{person}{Andreas Krause}, \bibinfo{person}{Emma Brunskill}, \bibinfo{person}{Kyunghyun Cho}, \bibinfo{person}{Barbara Engelhardt}, \bibinfo{person}{Sivan Sabato}, {and} \bibinfo{person}{Jonathan Scarlett}} (Eds.). \bibinfo{publisher}{{PMLR}}, \bibinfo{pages}{17283--17300}.
\newblock
\urldef\tempurl%
\url{https://proceedings.mlr.press/v202/koh23a.html}
\showURL{%
\tempurl}


\bibitem[Lerer et~al\mbox{.}(2019)]%
        {pbg}
\bibfield{author}{\bibinfo{person}{Adam Lerer}, \bibinfo{person}{Ledell Wu}, \bibinfo{person}{Jiajun Shen}, \bibinfo{person}{Timothee Lacroix}, \bibinfo{person}{Luca Wehrstedt}, \bibinfo{person}{Abhijit Bose}, {and} \bibinfo{person}{Alex Peysakhovich}.} \bibinfo{year}{2019}\natexlab{}.
\newblock \showarticletitle{{PyTorch-BigGraph: A Large-scale Graph Embedding System}}. In \bibinfo{booktitle}{\emph{Proceedings of the 2nd SysML Conference}}. \bibinfo{address}{Palo Alto, CA, USA}.
\newblock


\bibitem[Lewis et~al\mbox{.}(2020)]%
        {Lewis2020}
\bibfield{author}{\bibinfo{person}{Patrick S.~H. Lewis}, \bibinfo{person}{Ethan Perez}, \bibinfo{person}{Aleksandra Piktus}, \bibinfo{person}{Fabio Petroni}, \bibinfo{person}{Vladimir Karpukhin}, \bibinfo{person}{Naman Goyal}, \bibinfo{person}{Heinrich K{\"{u}}ttler}, \bibinfo{person}{Mike Lewis}, \bibinfo{person}{Wen{-}tau Yih}, \bibinfo{person}{Tim Rockt{\"{a}}schel}, \bibinfo{person}{Sebastian Riedel}, {and} \bibinfo{person}{Douwe Kiela}.} \bibinfo{year}{2020}\natexlab{}.
\newblock \showarticletitle{Retrieval-Augmented Generation for Knowledge-Intensive {NLP} Tasks}. In \bibinfo{booktitle}{\emph{Advances in Neural Information Processing Systems 33: Annual Conference on Neural Information Processing Systems 2020, NeurIPS 2020, December 6-12, 2020, virtual}}, \bibfield{editor}{\bibinfo{person}{Hugo Larochelle}, \bibinfo{person}{Marc'Aurelio Ranzato}, \bibinfo{person}{Raia Hadsell}, \bibinfo{person}{Maria{-}Florina Balcan}, {and} \bibinfo{person}{Hsuan{-}Tien Lin}} (Eds.).
\newblock


\bibitem[Li et~al\mbox{.}(2023)]%
        {HaluEval}
\bibfield{author}{\bibinfo{person}{Junyi Li}, \bibinfo{person}{Xiaoxue Cheng}, \bibinfo{person}{Wayne~Xin Zhao}, \bibinfo{person}{Jian-Yun Nie}, {and} \bibinfo{person}{Ji-Rong Wen}.} \bibinfo{year}{2023}\natexlab{}.
\newblock \bibinfo{title}{HaluEval: A Large-Scale Hallucination Evaluation Benchmark for Large Language Models}.
\newblock
\newblock
\urldef\tempurl%
\url{https://arxiv.org/abs/2305.11747}
\showURL{%
\tempurl}


\bibitem[Li et~al\mbox{.}(2024)]%
        {li2024topa}
\bibfield{author}{\bibinfo{person}{Wei Li}, \bibinfo{person}{Hehe Fan}, \bibinfo{person}{Yongkang Wong}, \bibinfo{person}{Mohan Kankanhalli}, {and} \bibinfo{person}{Yi Yang}.} \bibinfo{year}{2024}\natexlab{}.
\newblock \bibinfo{title}{TOPA: Extend Large Language Models for Video Understanding via Text-Only Pre-Alignment}.
\newblock
\newblock
\showeprint[arxiv]{2405.13911}~[cs.CV]
\urldef\tempurl%
\url{https://arxiv.org/abs/2405.13911}
\showURL{%
\tempurl}


\bibitem[Lin et~al\mbox{.}(2023)]%
        {Lin2023}
\bibfield{author}{\bibinfo{person}{Bin Lin}, \bibinfo{person}{Yang Ye}, \bibinfo{person}{Bin Zhu}, \bibinfo{person}{Jiaxi Cui}, \bibinfo{person}{Munan Ning}, \bibinfo{person}{Peng Jin}, {and} \bibinfo{person}{Li Yuan}.} \bibinfo{year}{2023}\natexlab{}.
\newblock \showarticletitle{Video-LLaVA: Learning United Visual Representation by Alignment Before Projection}.
\newblock \bibinfo{journal}{\emph{CoRR}}  \bibinfo{volume}{abs/2311.10122} (\bibinfo{year}{2023}).
\newblock
\urldef\tempurl%
\url{https://doi.org/10.48550/ARXIV.2311.10122}
\showDOI{\tempurl}
\showeprint[arXiv]{2311.10122}


\bibitem[Lin et~al\mbox{.}(2022)]%
        {lin-etal-2022-truthfulqa}
\bibfield{author}{\bibinfo{person}{Stephanie Lin}, \bibinfo{person}{Jacob Hilton}, {and} \bibinfo{person}{Owain Evans}.} \bibinfo{year}{2022}\natexlab{}.
\newblock \showarticletitle{{T}ruthful{QA}: Measuring How Models Mimic Human Falsehoods}. In \bibinfo{booktitle}{\emph{Proceedings of the 60th Annual Meeting of the Association for Computational Linguistics (Volume 1: Long Papers)}}, \bibfield{editor}{\bibinfo{person}{Smaranda Muresan}, \bibinfo{person}{Preslav Nakov}, {and} \bibinfo{person}{Aline Villavicencio}} (Eds.). \bibinfo{publisher}{Association for Computational Linguistics}, \bibinfo{address}{Dublin, Ireland}, \bibinfo{pages}{3214--3252}.
\newblock
\urldef\tempurl%
\url{https://doi.org/10.18653/v1/2022.acl-long.229}
\showDOI{\tempurl}


\bibitem[Liu et~al\mbox{.}(2024)]%
        {liu2024improved}
\bibfield{author}{\bibinfo{person}{Haotian Liu}, \bibinfo{person}{Chunyuan Li}, \bibinfo{person}{Yuheng Li}, {and} \bibinfo{person}{Yong~Jae Lee}.} \bibinfo{year}{2024}\natexlab{}.
\newblock \bibinfo{title}{Improved Baselines with Visual Instruction Tuning}.
\newblock
\newblock
\showeprint[arxiv]{2310.03744}~[cs.CV]
\urldef\tempurl%
\url{https://arxiv.org/abs/2310.03744}
\showURL{%
\tempurl}


\bibitem[Liu et~al\mbox{.}(2023)]%
        {LiuLWL23a}
\bibfield{author}{\bibinfo{person}{Haotian Liu}, \bibinfo{person}{Chunyuan Li}, \bibinfo{person}{Qingyang Wu}, {and} \bibinfo{person}{Yong~Jae Lee}.} \bibinfo{year}{2023}\natexlab{}.
\newblock \showarticletitle{Visual Instruction Tuning}. In \bibinfo{booktitle}{\emph{Advances in Neural Information Processing Systems 36: Annual Conference on Neural Information Processing Systems 2023, NeurIPS 2023, New Orleans, LA, USA, December 10 - 16, 2023}}, \bibfield{editor}{\bibinfo{person}{Alice Oh}, \bibinfo{person}{Tristan Naumann}, \bibinfo{person}{Amir Globerson}, \bibinfo{person}{Kate Saenko}, \bibinfo{person}{Moritz Hardt}, {and} \bibinfo{person}{Sergey Levine}} (Eds.).
\newblock
\urldef\tempurl%
\url{http://papers.nips.cc/paper\_files/paper/2023/hash/6dcf277ea32ce3288914faf369fe6de0-Abstract-Conference.html}
\showURL{%
\tempurl}


\bibitem[Liu et~al\mbox{.}(2019)]%
        {liu2019roberta}
\bibfield{author}{\bibinfo{person}{Yinhan Liu}, \bibinfo{person}{Myle Ott}, \bibinfo{person}{Naman Goyal}, \bibinfo{person}{Jingfei Du}, \bibinfo{person}{Mandar Joshi}, \bibinfo{person}{Danqi Chen}, \bibinfo{person}{Omer Levy}, \bibinfo{person}{Mike Lewis}, \bibinfo{person}{Luke Zettlemoyer}, {and} \bibinfo{person}{Veselin Stoyanov}.} \bibinfo{year}{2019}\natexlab{}.
\newblock \showarticletitle{Roberta: A robustly optimized bert pretraining approach}.
\newblock \bibinfo{journal}{\emph{arXiv preprint arXiv:1907.11692}} (\bibinfo{year}{2019}).
\newblock


\bibitem[Moon et~al\mbox{.}(2019)]%
        {moon-etal-2019-opendialkg}
\bibfield{author}{\bibinfo{person}{Seungwhan Moon}, \bibinfo{person}{Pararth Shah}, \bibinfo{person}{Anuj Kumar}, {and} \bibinfo{person}{Rajen Subba}.} \bibinfo{year}{2019}\natexlab{}.
\newblock \showarticletitle{{O}pen{D}ial{KG}: Explainable Conversational Reasoning with Attention-based Walks over Knowledge Graphs}. In \bibinfo{booktitle}{\emph{Proceedings of the 57th Annual Meeting of the Association for Computational Linguistics}}, \bibfield{editor}{\bibinfo{person}{Anna Korhonen}, \bibinfo{person}{David Traum}, {and} \bibinfo{person}{Llu{\'\i}s M{\`a}rquez}} (Eds.). \bibinfo{publisher}{Association for Computational Linguistics}, \bibinfo{address}{Florence, Italy}, \bibinfo{pages}{845--854}.
\newblock
\urldef\tempurl%
\url{https://doi.org/10.18653/v1/P19-1081}
\showDOI{\tempurl}


\bibitem[Ouyang et~al\mbox{.}(2022)]%
        {ouyang2022training}
\bibfield{author}{\bibinfo{person}{Long Ouyang}, \bibinfo{person}{Jeff Wu}, \bibinfo{person}{Xu Jiang}, \bibinfo{person}{Diogo Almeida}, \bibinfo{person}{Carroll~L. Wainwright}, \bibinfo{person}{Pamela Mishkin}, \bibinfo{person}{Chong Zhang}, \bibinfo{person}{Sandhini Agarwal}, \bibinfo{person}{Katarina Slama}, \bibinfo{person}{Alex Ray}, \bibinfo{person}{John Schulman}, \bibinfo{person}{Jacob Hilton}, \bibinfo{person}{Fraser Kelton}, \bibinfo{person}{Luke Miller}, \bibinfo{person}{Maddie Simens}, \bibinfo{person}{Amanda Askell}, \bibinfo{person}{Peter Welinder}, \bibinfo{person}{Paul Christiano}, \bibinfo{person}{Jan Leike}, {and} \bibinfo{person}{Ryan Lowe}.} \bibinfo{year}{2022}\natexlab{}.
\newblock \bibinfo{title}{Training language models to follow instructions with human feedback}.
\newblock
\newblock
\showeprint[arxiv]{2203.02155}~[cs.CL]
\urldef\tempurl%
\url{https://arxiv.org/abs/2203.02155}
\showURL{%
\tempurl}


\bibitem[Sakaguchi et~al\mbox{.}(2021)]%
        {sakaguchi2021winogrande}
\bibfield{author}{\bibinfo{person}{Keisuke Sakaguchi}, \bibinfo{person}{Ronan~Le Bras}, \bibinfo{person}{Chandra Bhagavatula}, {and} \bibinfo{person}{Yejin Choi}.} \bibinfo{year}{2021}\natexlab{}.
\newblock \showarticletitle{Winogrande: An adversarial winograd schema challenge at scale}.
\newblock \bibinfo{journal}{\emph{Commun. ACM}} \bibinfo{volume}{64}, \bibinfo{number}{9} (\bibinfo{year}{2021}), \bibinfo{pages}{99--106}.
\newblock


\bibitem[See et~al\mbox{.}(2017)]%
        {see-etal-2017-get}
\bibfield{author}{\bibinfo{person}{Abigail See}, \bibinfo{person}{Peter~J. Liu}, {and} \bibinfo{person}{Christopher~D. Manning}.} \bibinfo{year}{2017}\natexlab{}.
\newblock \showarticletitle{Get To The Point: Summarization with Pointer-Generator Networks}. In \bibinfo{booktitle}{\emph{Proceedings of the 55th Annual Meeting of the Association for Computational Linguistics (Volume 1: Long Papers)}}, \bibfield{editor}{\bibinfo{person}{Regina Barzilay} {and} \bibinfo{person}{Min-Yen Kan}} (Eds.). \bibinfo{publisher}{Association for Computational Linguistics}, \bibinfo{address}{Vancouver, Canada}, \bibinfo{pages}{1073--1083}.
\newblock
\urldef\tempurl%
\url{https://doi.org/10.18653/v1/P17-1099}
\showDOI{\tempurl}


\bibitem[Sun et~al\mbox{.}(2023)]%
        {sun-etal-2023-towards}
\bibfield{author}{\bibinfo{person}{Bin Sun}, \bibinfo{person}{Yitong Li}, \bibinfo{person}{Fei Mi}, \bibinfo{person}{Fanhu Bie}, \bibinfo{person}{Yiwei Li}, {and} \bibinfo{person}{Kan Li}.} \bibinfo{year}{2023}\natexlab{}.
\newblock \showarticletitle{Towards Fewer Hallucinations in Knowledge-Grounded Dialogue Generation via Augmentative and Contrastive Knowledge-Dialogue}. In \bibinfo{booktitle}{\emph{Proceedings of the 61st Annual Meeting of the Association for Computational Linguistics (Volume 2: Short Papers)}}, \bibfield{editor}{\bibinfo{person}{Anna Rogers}, \bibinfo{person}{Jordan Boyd-Graber}, {and} \bibinfo{person}{Naoaki Okazaki}} (Eds.). \bibinfo{publisher}{Association for Computational Linguistics}, \bibinfo{address}{Toronto, Canada}, \bibinfo{pages}{1741--1750}.
\newblock
\urldef\tempurl%
\url{https://doi.org/10.18653/v1/2023.acl-short.148}
\showDOI{\tempurl}


\bibitem[Tang et~al\mbox{.}(2023)]%
        {Tang2023}
\bibfield{author}{\bibinfo{person}{Changli Tang}, \bibinfo{person}{Wenyi Yu}, \bibinfo{person}{Guangzhi Sun}, \bibinfo{person}{Xianzhao Chen}, \bibinfo{person}{Tian Tan}, \bibinfo{person}{Wei Li}, \bibinfo{person}{Lu Lu}, \bibinfo{person}{Zejun Ma}, {and} \bibinfo{person}{Chao Zhang}.} \bibinfo{year}{2023}\natexlab{}.
\newblock \showarticletitle{{SALMONN:} Towards Generic Hearing Abilities for Large Language Models}.
\newblock \bibinfo{journal}{\emph{CoRR}}  \bibinfo{volume}{abs/2310.13289} (\bibinfo{year}{2023}).
\newblock
\urldef\tempurl%
\url{https://doi.org/10.48550/ARXIV.2310.13289}
\showDOI{\tempurl}
\showeprint[arXiv]{2310.13289}


\bibitem[Thorne et~al\mbox{.}(2018)]%
        {Thorne18Fever}
\bibfield{author}{\bibinfo{person}{James Thorne}, \bibinfo{person}{Andreas Vlachos}, \bibinfo{person}{Christos Christodoulopoulos}, {and} \bibinfo{person}{Arpit Mittal}.} \bibinfo{year}{2018}\natexlab{}.
\newblock \showarticletitle{{FEVER}: a Large-scale Dataset for Fact Extraction and {VERification}}. In \bibinfo{booktitle}{\emph{NAACL-HLT}}.
\newblock


\bibitem[Touvron et~al\mbox{.}(2023)]%
        {touvron2023llama}
\bibfield{author}{\bibinfo{person}{Hugo Touvron}, \bibinfo{person}{Thibaut Lavril}, \bibinfo{person}{Gautier Izacard}, \bibinfo{person}{Xavier Martinet}, \bibinfo{person}{Marie-Anne Lachaux}, \bibinfo{person}{Timothée Lacroix}, \bibinfo{person}{Baptiste Rozière}, \bibinfo{person}{Naman Goyal}, \bibinfo{person}{Eric Hambro}, \bibinfo{person}{Faisal Azhar}, \bibinfo{person}{Aurelien Rodriguez}, \bibinfo{person}{Armand Joulin}, \bibinfo{person}{Edouard Grave}, {and} \bibinfo{person}{Guillaume Lample}.} \bibinfo{year}{2023}\natexlab{}.
\newblock \bibinfo{title}{LLaMA: Open and Efficient Foundation Language Models}.
\newblock
\newblock
\showeprint[arxiv]{2302.13971}~[cs.CL]


\bibitem[Vrande\v{c}i\'{c} and Kr\"{o}tzsch(2014)]%
        {10.1145/2629489}
\bibfield{author}{\bibinfo{person}{Denny Vrande\v{c}i\'{c}} {and} \bibinfo{person}{Markus Kr\"{o}tzsch}.} \bibinfo{year}{2014}\natexlab{}.
\newblock \showarticletitle{Wikidata: a free collaborative knowledgebase}.
\newblock \bibinfo{journal}{\emph{Commun. ACM}} \bibinfo{volume}{57}, \bibinfo{number}{10} (\bibinfo{date}{sep} \bibinfo{year}{2014}), \bibinfo{pages}{78–85}.
\newblock
\showISSN{0001-0782}
\urldef\tempurl%
\url{https://doi.org/10.1145/2629489}
\showDOI{\tempurl}


\bibitem[Xu et~al\mbox{.}(2023)]%
        {xu-etal-2023-fine}
\bibfield{author}{\bibinfo{person}{Silei Xu}, \bibinfo{person}{Shicheng Liu}, \bibinfo{person}{Theo Culhane}, \bibinfo{person}{Elizaveta Pertseva}, \bibinfo{person}{Meng-Hsi Wu}, \bibinfo{person}{Sina Semnani}, {and} \bibinfo{person}{Monica Lam}.} \bibinfo{year}{2023}\natexlab{}.
\newblock \showarticletitle{Fine-tuned {LLM}s Know More, Hallucinate Less with Few-Shot Sequence-to-Sequence Semantic Parsing over {W}ikidata}. In \bibinfo{booktitle}{\emph{Proceedings of the 2023 Conference on Empirical Methods in Natural Language Processing}}, \bibfield{editor}{\bibinfo{person}{Houda Bouamor}, \bibinfo{person}{Juan Pino}, {and} \bibinfo{person}{Kalika Bali}} (Eds.). \bibinfo{publisher}{Association for Computational Linguistics}, \bibinfo{address}{Singapore}, \bibinfo{pages}{5778--5791}.
\newblock
\urldef\tempurl%
\url{https://doi.org/10.18653/v1/2023.emnlp-main.353}
\showDOI{\tempurl}


\bibitem[Yang et~al\mbox{.}(2018)]%
        {yang-etal-2018-hotpotqa}
\bibfield{author}{\bibinfo{person}{Zhilin Yang}, \bibinfo{person}{Peng Qi}, \bibinfo{person}{Saizheng Zhang}, \bibinfo{person}{Yoshua Bengio}, \bibinfo{person}{William Cohen}, \bibinfo{person}{Ruslan Salakhutdinov}, {and} \bibinfo{person}{Christopher~D. Manning}.} \bibinfo{year}{2018}\natexlab{}.
\newblock \showarticletitle{{H}otpot{QA}: A Dataset for Diverse, Explainable Multi-hop Question Answering}. In \bibinfo{booktitle}{\emph{Proceedings of the 2018 Conference on Empirical Methods in Natural Language Processing}}, \bibfield{editor}{\bibinfo{person}{Ellen Riloff}, \bibinfo{person}{David Chiang}, \bibinfo{person}{Julia Hockenmaier}, {and} \bibinfo{person}{Jun{'}ichi Tsujii}} (Eds.). \bibinfo{publisher}{Association for Computational Linguistics}, \bibinfo{address}{Brussels, Belgium}, \bibinfo{pages}{2369--2380}.
\newblock
\urldef\tempurl%
\url{https://doi.org/10.18653/v1/D18-1259}
\showDOI{\tempurl}


\bibitem[Zellers et~al\mbox{.}(2019)]%
        {zellers2019hellaswag}
\bibfield{author}{\bibinfo{person}{Rowan Zellers}, \bibinfo{person}{Ari Holtzman}, \bibinfo{person}{Yonatan Bisk}, \bibinfo{person}{Ali Farhadi}, {and} \bibinfo{person}{Yejin Choi}.} \bibinfo{year}{2019}\natexlab{}.
\newblock \showarticletitle{HellaSwag: Can a Machine Really Finish Your Sentence?}
\newblock \bibinfo{journal}{\emph{CoRR}}  \bibinfo{volume}{abs/1905.07830} (\bibinfo{year}{2019}).
\newblock
\showeprint[arXiv]{1905.07830}
\urldef\tempurl%
\url{http://arxiv.org/abs/1905.07830}
\showURL{%
\tempurl}


\bibitem[Zhong et~al\mbox{.}(2024)]%
        {zhong2024enhancing}
\bibfield{author}{\bibinfo{person}{Wenliang Zhong}, \bibinfo{person}{Wenyi Wu}, \bibinfo{person}{Qi Li}, \bibinfo{person}{Rob Barton}, \bibinfo{person}{Boxin Du}, \bibinfo{person}{Shioulin Sam}, \bibinfo{person}{Karim Bouyarmane}, \bibinfo{person}{Ismail Tutar}, {and} \bibinfo{person}{Junzhou Huang}.} \bibinfo{year}{2024}\natexlab{}.
\newblock \bibinfo{title}{Enhancing Multimodal Large Language Models with Multi-instance Visual Prompt Generator for Visual Representation Enrichment}.
\newblock
\newblock
\showeprint[arxiv]{2406.02987}~[cs.CV]
\urldef\tempurl%
\url{https://arxiv.org/abs/2406.02987}
\showURL{%
\tempurl}


\end{thebibliography}

\appendix

\appendix
\onecolumn

\section{Appendix A}
\label{sec:appendixA}
\vspace{2em}

\begin{tcolorbox}[colframe=black, colback=white, boxrule=0.5mm, arc=2mm]
\vspace{1em}
I want you act as an statement judge. Given a statement, your objective is to determine if the provided statement correct or not. Write "True" if the given statement is true and "False" if given statement is false. The answer you give MUST be "True" or "False".\\

For example:\\

Statement: Marlon Brando refused to be in any films in 1972.

Your Judgement: "False"\\

Statement: Dracula is a written work.

Your Judgement: "True"\\

Statement: Maria Sharapova passed all drug tests at the 2016 Australian Open.

Your Judgement: "False"\\

Statement: Blood are red.

Your Judgement: "True".\\

Statement: David Cronenberg has a cameo in To Die For.

Your Judgement: "True".\\

Statement: Albert Einstein was born in Moscow.

Your Judgement: "False".\\

Statement: Cubango is a city in Australia.

Your Judgement: "False".\\

Statement: The Okavango River is a river in southwest Africa.

Your Judgement: "True".\\


\end{tcolorbox}

\captionsetup{type=figure}
\captionof{figure}{The default 8-shot prompt for True-False benchmark.}

\vspace{2em}

\begin{tcolorbox}[colframe=black, colback=white, boxrule=0.5mm, arc=2mm]
\vspace{1em}
I want you act as an statement judge. Given a statement, your objective is to determine if the provided statement correct or not. Write "SUPPORTS" if the given statement is true and "REFUTES" if given statement is false. The answer you give MUST be "SUPPORTS" or "REFUTES". \\

For example: \\

Statement: Furia is adapted from Graffiti and it is French.\\
Your Judgement: "SUPPORTS" \\

Statement: Prince (musician) was not backed by the New Power Generation.\\
Your Judgement: "REFUTES" \\

Statement: Ronin is a 2001 film.\\
Your Judgement: "REFUTES" \\

Statement: The 1998 NFL Draft was cancelled April 18 -- 19.\\
Your Judgement: "REFUTES" \\

Statement: Usher's sophomore single is My Way.\\
Your Judgement: "REFUTES" \\

Statement: Anna Kendrick made her film debut in Camp.\\
Your Judgement: "SUPPORTS" \\

Statement: The G1 Climax is held each March.\\
Your Judgement: "REFUTES" \\

Statement: Loving was directed by Michael Shannon.\\
Your Judgement: "REFUTES" \\


\end{tcolorbox}

\captionsetup{type=figure}
\captionof{figure}{The default 8-shot prompt for FEVER dataset.}

\end{document}